\documentclass[12pt]{article}
\usepackage{a4}
\usepackage{dashbox}
\usepackage{scalerel,stackengine} % for widehats
\usepackage{expex}   % for all numbered examples 
\lingset{
    belowglpreambleskip=-0.2ex,% shrinks the vertical space between the preamble and the top gloss line
    everygla=,% removes the default italic formatting of the top gloss line
    aboveglftskip=-0.2ex,% shrinks the vertical space between the aligned lines and the free translation line
    interpartskip=0ex,% vertical space between parts of examples
    glspace=!0pt plus .2em,% improves line breaking by increasing the maximum horizontal space between aligned words
    glrightskip=0pt plus .5\hsize,% improves line breaking by increasing the maximum horizontal space between end of line and the right margin
    aboveexskip= .4ex plus .4ex minus .4ex,% vertical space above examples
    belowexskip=.4ex plus .4ex minus .4ex,% vertical space below examples\
    labelalign=center,                     % looks better in large labels
    exnotype=arabic,               % easier to remember and refer
    numoffset=0pt,  % left justified example numbers
    textoffset=.4ex,  % we need the extra space we get by small label distance
    labeloffset=2pt
}

\usepackage{mathptmx}  % this book's font
\usepackage{eucal}     % better \cal font--optional
\usepackage{ccg-latex} % ALL CCG notation here. all syncat are \cat{..}. 
\usepackage{sm-latex}  % the S_M machine tree drawings. we can do trees.
\usepackage{letterspace,setspace,url}

\usepackage{datetime}
\usepackage{qtree}     % no more pdf tress
\usepackage{latexsym,amssymb,amsmath,stmaryrd} % need some symbols from here
\usepackage{bm}  % times have trouble with bold greek math
\usepackage{textcomp}
\usepackage[T1]{tipa}
\usepackage{ipa}
\usepackage[icelandic,english]{babel}        
\usepackage[T1]{fontenc}                      
\usepackage{natbib}
\usepackage{abbrevs}
\makeatother
\usepackage[frame,graph,arrow]{xy}
\makeatletter

\newcommand{\dotarrow}{% to be used in math mode...
   \mathrel{\ooalign{\hss\raise.65ex\hbox{\scalebox{1.2}{\normalfont .}}%
   \kern0.35ex\hss\cr$\rightarrow$}}}
\newcommand{\bfeats}[1]{\mbox{\tiny\framebox(9,5)[b]{#1}}}
\newcommand{\bfeat}[1]{\mbox{\tiny\framebox(13,5)[b]{#1}}} % boxed feature
\newcommand{\cgdotline}[2]              % writing a dotted CG line. 2 arguments
 {\mc{#1}{\dotfill \raisebox{-.30ex}{\tiny #2}}}% #1=# col #2=rule type
\newcommand{\pagodasize}{\footnotesize}    % default size for S_m trees, yes they look like pagodas
\newcommand{\spagodasize}{\scriptsize}      % smaller ones  
\renewcommand{\pagodasize}{\spagodasize} % to try everything in one size
\newcommand{\xref}[1]{~(\ref{#1})}   % for local reference (no chapter number). control space before example cite.

\newcommand{\lbr}{\ensuremath{[}}
\newcommand{\rbr}{\ensuremath{]}}

\newcommand{\xg}[1]{\mbox{\textsc{\MakeLowercase{#1}}}} % small caps in ex. glosses
\makeatletter
\newcommand*{\textoverline}[1]{$\overline{\hbox{#1}}\m@th$}
\makeatother

\newsavebox\foobox
\def\slantvalue{0}
\newcommand{\slantboxengine}[2][\slantvalue]{\mbox{%
        \sbox{\foobox}{#2}%
        \hskip\wd\foobox
        \pdfsave
        \pdfsetmatrix{1 0 #1 1}%
        \llap{\usebox{\foobox}}%
        \pdfrestore
}}
\newcommand\slantbox[2][\slantvalue]{%
  \edef\slantvalue{#1}\expandafter\slantboxhelpA#2 \relax\relax}
\def\slantboxhelpA#1 #2\relax{%
  \slantboxengine{#1}%
  \ifx\relax#2\relax\else\ \slantboxhelpA#2\relax\fi
}
\newcommand{\tool}{\mbox{\textsc{TheBench}}}

\usepackage{color,changes,verbatim}

\begin{document}
\thispagestyle{empty}
\newcommand{\keep}{\added}
\centerline{{\Large\textbf{{\tool~}}}~~~\texttt{\Large Guide}\hfill\hfill{\footnotesize Cem Boz\c{s}ahin$\mid$Ankara, Dat\c{c}a, \c{S}ile 2022--24}}

\noindent For Linguistic Analysis, Description\hfill\texttt{\footnotesize cem.bozsahin@gmail.com}\\ and Typological Exploration with Categorial Grammar\medskip\medskip

\noindent\keep{Version: 2.0~~~{\small\today}\hfill coloured text means changes from the previous release}\\
Home: \verb|github.com/bozsahin/thebench|~~~{\footnotesize(there are instructions here for install and use)}\medskip

\section{Introduction}
\tool\, is a tool to study monadic structures in natural language. It is for writing monadic grammars to explore analyses, compare diverse languages through their categories, and to train  models of grammar from form-meaning pairs where syntax is latent variable.
%One such grammar is described in \cite{bozs:mons}, hereafter \mg.

Monadic structures are binary combinations of elements that employ semantics of composition only, in a hermetic seal, hence the name; see \cite{macl:71,mogg:88} for that. \tool\,is essentially old-school categorial grammar to syntacticize the idea, with the implication that although syntax is autonomous  (recall \emph{colorless green ideas sleep furiously}), the treasure
is in the baggage it carries at every step, viz. semantics, more narrowly, predicate-argument structures indicating choice of categorial reference
and its consequent placeholders for decision in such structures. 

There is some new thought in old school.
Unlike traditional categorial grammars, application is turned into composition in monadic analysis. Moreover,
every correspondence requires specifying two command relations, one on syntactic command and the other on semantic command.
A monadic grammar of \tool\,
contains only synthetic elements (called `objects' in category theory of mathematics) that are shaped by this analytic invariant, viz. composition. 
Both ingredients (command relations) of any analytic step must therefore be functions (`arrows' in category theory). \tool\,is one implementation of the idea for iterative development of such 
functions along with grammar of synthetic elements.

Having to have functions  requires a bit of explaining on the linguistic side, which I do in the next section. If you are  more interested in the software specification and tool's use, please skip to \S\ref{sec:org}.

\section{Linguistics and mathematics behind the tool}
Mathematically, a monad composes two functions $f$ and $g$ to maintain the dependency of $f$ on $g$, which we can write in extensional terms as $\lambda x.f(gx)$. 

It does so
by doing $g \circ f$.  Computationally, it first gets $g$, then $f$.
The ultimate element $f$ is always the head function; see \citealt{gall:11}:118 for  reasons for the notation. In this representation, $g \circ f$,
it is clearer that $f$ {depends} on $g$ in $\lambda x.f(gx)$, not the other way around. In the alternative notation, writing instead $f \circ g$  to mean $\lambda x.f(gx)$, and saying that $f \circ g$ means
`do $g$, {then} $f$' to get the dependency right, is somewhat counter-intutitive, especially if we are going to construct all structure from sequencing and reference categorization by a single chain of linguistic dependency. Note that
$\lambda x.f(gx)$ dependency is a single chain: $g$ depends only on $x$, and $f$ depends only on $(gx)$.

Linguistically, the natural language monad of \tool\,internally turns $f$ and $g$ into semantic functions even if one of them is syntactically not a function. 
(If none of them is a  \emph{syntactic} function, they won't combine, with implications---in my opinion---for morphological compounding.)
It does so by deriving case functions from the verbs and verb-like elements of a grammar, and by keeping
intact two command relations, which every element of grammar specifies: surface command (s-command) and predicate-argument command (l-command).

S-command specifies in what order and directionality an element takes its syntactic arguments. It is asymmetrically structural, not a linear metric,
because it embodies the syntactic notion of combining later or earlier than other elements.
L-command specifies in what order and dominance the predicate-argument structure is revealed. It is similarly asymmetrically structural.
Pairing of the two is always required in the monadic grammar of \tool. %This is not a formal requirement; it is based on the idea that choice determines reference, and reference needs access to both syntax and semantics to determine surface structure  via first-order linguistic categorization subject to configurational constraints.

This pairing, in the form of a 
\emph{procedural category}, is a concept we can think of as a culmination of all of the following in one representational   package for  singular computation: Montague's (\citeyear{Montague73}) basic rules associating every syntactic application with logical interpretation,  Bach's (\citeyear{bach:76}) 
rule-to-rule thesis, Klein and Sag's (\citeyear{Klei:85}) one-way translation of syntactic and semantic  types, 
Grimshaw's (\citeyear{grimshaw90}) a-structure, and Williams's (\citeyear{Will:94}) argument complex, without nonlocality, that is, without
specification of argument's arguments---maintaining his locality of predication, and Hale and Keyser's (\citeyear{halekeyser02}) syntactic
configuration of a lexical item and its semantic components. 

A procedural category has 
 the added role of choice of reference \emph{for the whole} reflected in its syntactic type to lay out decision points for truth conditions of the parts in the semantic type.
The main implication for the grammarian is
that, perforce, there is a need for  language-variant linguistic vocabulary  to explore case, agreement, grammatical relations and other structural functions given such invariant analytics, because only
 variant vocabulary and preserved command relations seem to be able to make \emph{any grammar} transparent and consistently
 interpretable with respect to invariance.
 
Syntactic, phonological, morphological and predicate-argument structural reflex of composition-only analytics is the main subject of studying monadic structures in natural language. \tool\,is a tool implementing the idea. To sum up,
it is an old-school categorial grammar, except that it uses composition only. That is, application is also turned into composition, and, multiple dependency projection in one step (such as CCG's \combs\,projection) is not analytic; it has to be specified by the head of a construction.

Turning application into composition works as follows. Consider the following analysis in the context of applicative categorial grammar, much like \cite{ajdu:35,BarH:53,Montague73}:

\ex
{\pagodasize
\begin{smex}
  \smres{\cat{S}\\ \lf{\so{admire}\so{john}\so{sincerity}}}{}{4em}{2em}\\
  \smleaf{\cat{\cgs{NP}{3s}}\\ \lf{\so{sincerity}}\\ \phon{Sincerity}} 
                   & \begin{smex}
                       \smresl{\cat{S\bs\cgs{NP}{3s}}\\ \lf{\lambda y.\so{admire}\so{john}y}}{}{4em}\\
                       \smleaf{\cat{(S\bs\cgs{NP}{3s})\fs NP}\\ \lf{\lambda x\lambda y.\so{admire}xy}\\ \phon{admires}} &
                       \smleaf{\cat{NP}\\ \lf{\so{john}}\\ \phon{John}}
                    \end{smex}
\end{smex}
}
\xe

Here, the verb applies to its arguments one at a time, first to its syntactic
object, then to its subject. Application is evident on the semantic side, written after `:'.

What takes places inside the monad is the following, where both applications are via function composition, in the template of $\lambda z.f(gz)$. And, the $f$ function, the head,  is always the ultimate element:

\ex\label{ex:mdemo}
{\pagodasize
\begin{smex}
  \smres{\cat{S}\\ \lf{\so{admire}\so{john}\so{sincerity}=}\\
                            \blf{\lambda z.\lbr\underline{\lambda \psi.\psi(\lambda y.\so{admire}\so{john}y)}\rbr\lbr(\lambda \phi.\phi\so{sincerity})z\rbr}}{}{7em}{4em}\\
  \smleaf{\cat{\cgs{NP}{3s}}\\ \lf{\so{sincerity}\nearrow\lambda \phi.\phi\so{sincerity}}\\ \phon{Sincerity}} 
                   & \hspace*{-6em}\begin{smex}
                       \smresl{\cat{S\bs\cgs{NP}{3s}}\\ \lf{\lambda y.\so{admire}\so{john}\,y=}\\ 
                       \blf{\lambda z.\lbr\underline{\lambda \psi.\psi\,\so{john}}\rbr\lbr(\lambda\phi.(\lambda x\lambda y.\so{admire}xy)\phi)z\rbr}\\
                       \blf{\nearrow \blf{{\lambda \psi.\psi(\lambda y.\so{admire}\so{john}y)}}}}{}{8em}\\
                       \smleaf{\cat{(S\bs\cgs{NP}{3s})\fs NP}\\ \lf{\lambda x\lambda y.\so{admire}xy\nearrow \lambda\phi.(\lambda x\lambda y.\so{admire}xy)\phi}\\ \phon{admires}} &
                       \smleaf{\cat{NP}\\ \lf{\so{john}\nearrow {\lambda \psi.\psi\,\so{john}}}\\ \phon{John}}
                    \end{smex}
\end{smex}
}
\xe
 A monad always composes this way. The head function, the one that determines the result, is always the ultimate function in the input to combination. It is underlined at every step in the preceding example.

The implication for linguistic theory is that, the semantic lifting that takes place to engender composition, which is represented in the example using $\nearrow$, must be accompanied by a lifted syntactic category, for all elements. For arguments of the verbs, these are the case functions, for any kind of argument including clausal arguments of \emph{think}-like verbs. Their narrowing in reference and choice can manifest itself for example in agreement. For example subject agreement reduces 
options for reference from \cat{S\fs(S\bs NP)} to \cat{S\fs(S\bs\cgs{NP}{3s})} for third-person singular subject, which can be read off the verbal subcategorization if distinct verb forms make the distinction in their categories. Cross-categorial generalizations are possible, also from verbal subcategorization, for example grammatical relations and systems of accusativity and ergativity. 

Notice that in the example the verb itself is also `lifted' semantically inside.
Although, formally speaking, its lifted type is eta-equivalent to its unlifted type, the implication is that the lifted type is
a function over and above who does what to whom.

For the verb itself, and for similarly complex categories, their lifting  can be associated empirically with spatiotemporality. Since $\phi$ in the example
is a transparent function arising from analytics, tense and aspect must be language-particular functions over verbal types (see \citealt{klei:94,klei:00}). For the arguments of verbs, the lifting of their basic syntactic type relates to case---because they are subcategorized for. Such functions are crosslinguistically inferrable in \tool\,from verbal subcategorizations; NB. the `\verb~c command~'.

For case, all  cases are structural (i.e. second-order) functions because they can be read off the first-order functions such as the verb. Identifying different subcategorizations of distinct verb forms (even for the same verb, as its root, stem and finite subcategorizations) leads to a bottom-up theory of `universals', that is,  categorizations available to everyone, from the ground up.

Unlike generative grammar, there are no top-down universals such as EPP, minimal link condition (MLC)  or universal lexical categories such as {N, V, A, P}, or universal argument types such as those in X-bar theory.  All of that is predicted from the ground up. The Husserlian and Polish-school idea is there,
that segments arise because analysis-by-categories seeks out reference fragments in an expression. We do not assume that analysis arises from (or rules `generate' from)  Harris-style distributional segments.

\tool\,is meant to explore configurationality, narrowly understood here as studying the limits on surface distributional  structure,
\emph{pace} \cite{Hale:83},\footnote{From this perspective, there is no such thing
as a non-configurational language in the sense of \cite{Hale:83}.}
and referentiality, together. \tool's configurational approach to surface structure is inspired by
 \cite{lambek58,stee:20}. Its referential approach to constructing elements is inspired by \cite{sapi:24,swad:38,Mont:70a,Montague73,schm:18}. The following aspects can help situate the proposal in the landscape of theories of grammar:
 
 Formally, as stated earlier, unlike categorial grammars such as that of \cite{steedman00}, all analytic structures handle one dependency at a time (i.e. there is no \combs-style analytic structure taking care
 of multiple dependencies at once). Unlike Montague grammar, predicate-argument structures are not post-revealed (\emph{de re, de dicto}, scope inversion, etc.), because different categorization due to different reference is expected
 to play the key role in the explanation.\footnote{Cf. \emph{every professor wrote a book} and \emph{every student missed a meal}.
 Not all predicates allow for inversion, therefore it is not a far-fetched idea to replace post-reveal of scope
 with differences in the referentiality of the predicate and consequent change in type-raising of the arguments.
 In other words, neither direct compositionality of \cite{Mont:70a} nor quantifying-in of \cite{Montague73} are suggested as alternatives, relying instead on
 the verb's referential properties and the category of the quantifiers and non-quantifiers. \tool\,is more compatible with the 1970 idea, but  arguing instead
 to support different subcategorization due to different referentiality.} Moreover, syntactic types
of second-order (structural)  functions including those for case, agreement and grammatical relations require a linguistic theory
precisely because they are all based on grounded (first order) elements.

Distinct function-argument and argument-function sequencing of reference is the cause of categorial and structural asymmetries in the monadic grammar of \tool, which is an idea that goes back to \cite{schonfinkel24}. He had used the idea
to motivate his combinators, all but three (binary \combb, \comba\,and \combt) considered to be too powerful in \tool\,for monadic structures. 
In their simplest forms they are:
\pex
\a \cat{X\fs Y}~~\cat{Y\fs Z} $\Rightarrow$ \cat{X\fs Z}\hfill (forward \combb)
\a \cat{Y\bs Z}~~\cat{X\bs Y} $\Rightarrow$ \cat{X\bs Z}\hfill (backward \combb)
\a \cat{X\fs Y}~~\cat{Y} $\Rightarrow$ \cat{X}\hfill (\comba)
\a \cat{Y}~~\cat{X\bs Y} $\Rightarrow$ \cat{X}\hfill (\combt)
\xe
\cite{steedmanbaldridge-guide} provide more information  about them.

What's left behind, composition including combination with \comba\, and \combt, is anybody's composition, 
but with some exploratory power arising from function-argument and argument-function distinction because it allows transparent transmission of reference.
We don't really know whether
choice is compositionally determined. We would be studying how far it can be transmitted compositionally in syntax starting with the whole, but not holistically.
(This sounds like a distant memory in psychology too; see \citealt{koff:36}).

Empirically, morphology and syntax do not compete for the same task in monadic analysis, for example for surface bracketing (morphology constructs and syntax transmits reference), 
morphology is not confined to the `lexicon', to `operators' or to some other component, or to leaves of a tree; it is an autonomous structure, obviously
not isomorphic to phonological or syntactic structure but also not subserving them either, projected
altogether homomorphically in analytic structure. Morphology is considered to be the 
category constructor for the form, therefore all languages have morphology. And, referential differences
in all kinds of arguments cause categorial differences including those in idioms and phrasal verbs \citep{bozs:22-jlli}.

Typologically, any language-particular  difference in elementary (i.e. synthetic) vocabulary can make its way into invariant analytic structure presuming  it is the empirical motive to make argument-taking transparent. It is a very verb-centric  view of grammar. Elements' morphology plays a crucial role in transparency of analytics. This is one reason we avoid commitment to certain ways of doing morphology in \tool.

\section{\tool\,organization}\label{sec:org}

Technically, editing functions of \tool, the functions that transform internal files to editable files, and interfaces of \tool, are all written in \verb~Python~. Processor functions are written in \textsc{Common Lisp}. 
The installer (NB. top of first page) takes care of the software requirements to work with all popular personal computer platforms (well, almost all).
The command
interface, which is given at the back, has its own syntax combining the two aspects for editing and processing, which you can use online and offline
(the `\verb~@ command~', aka. `batch mode'---see the basic glossary at the back). 

There is no need to know either \textsc{Common Lisp} or Python to use \tool. (Maybe this much is good to know to understand processor's output: \verb~T~ means true, and \verb~NIL~ means false in all Lisps.)
Use the `\verb~? command~' when you launch \tool\,to recall the full list of commands.
Figure~\ref{fig:welcome} shows what you see when you launch \tool. (Software versions may vary.)

\begin{figure}[t]
\footnotesize
\begin{verbatim}
-------------------------------------------------------------------------
Welcome to The Bench
    A workbench for studying NL structures built by two command relations
        Bench version:       2.0 Dated: March 25, 2024
        Python version:      3.10.10
        Common Lisp version: SBCL 2.2.9
 
        Pre/post processing by Python (grammar development, interfaces)
        Processing by Common Lisp     (analysis, training, ranking)
       
    Today: June 02, 2024, 12:41:36
Type x to exit, ? to get some help
-------------------------------------------------------------------------
lisp     : bench.lisp loaded, version 8.0, encoding UTF-8
python   : bench.py   loaded, version 2.0, encoding utf-8
ready
\end{verbatim}
\caption{Welcome screen of \tool\,(without \textsc{utf} codes for terminal prompts).}
\label{fig:welcome}
\end{figure}

Regarding organization of your files for work, the editable files of \tool, either written by you or generated by the tool, reside in your working directory.
Processor's internal files go into the directory created by \tool\, installer for you; they reside in \verb~/var/tmp/thebench~. 
Occasional clean-up of both locations is recommended.\footnote{See the end of \S\ref{sec:work} for when not to do a clean-up.}
Transformation  of an internal file to an editable grammar is possible,
which takes a file from \verb~/var/tmp/thebench~ to save the editable file in your working directory.

The training command `\verb~t~'  loads an extra processor file called
\verb~bench.user.lisp~ which is available in the repository. You can change that file, but please do not rename, relocate or delete it. 
If it is not loadable, the `\verb~t command~' would complain and exit.
No internal code
depends on the code contained in this file. It is useful for experimenting on model parameters without changing \tool\,code if you are going to do modeling.

Direct \textsc{ipa} support is too unwieldy for the tool; it seemed best to use \textsc{utf-8} tools for it in Python and Lisp.
When you launch \tool, please check the encoding reported by the processor. Both \verb~Python~ and
\textsc{Common Lisp} in your computer must report \textsc{utf-8}, 
as in Figure~\ref{fig:welcome}.
\section{Synthetics}
There are three kinds of elements of grammar in \tool:
\pex
\a  an elementary item,
\a an asymmetric relational rule, and
\a a symmetric relational rule.
\xe\smallskip

Examples in \tool\,notation are respectively: (margin overrun is deliberate)

\ex\label{ex:3el}
\footnotesize
\hspace*{-.7em}\begin{tabular}[t]{l}
 \verb~likes  | v :: (s\^np[agr=3s])/^np : \x\y.like x y     % ^: object can topicalize~\\
\verb~#np-raise np[agr=?x] : lf  --> s/(s\np[agr=?x]) : \lf\p. p lf  ~\\
\verb~#tense runs, s[t=pres,agr=3s]\np:\x.pres run x <--> ran, s[t=past]\np:\x.past run x~
\end{tabular}
\xe\vskip .5em

Every entry must start in a new line and it must be on one line, without line breaks. 
This allows us to minimize non-substantive punctuation. Let the longer lines wrap around on the screen. Everything starting with `\verb~%~' until the end of a line is considered to be a comment. Capitalization is significant only in 
phonological elements (the  material up to `\verb~|~'). 

Whitespacing is not important anywhere. Empty and comment-only lines are fine. The order of specification of elementary items and symmetric relational rules  in the grammar's text file is not important. Asymmetric relational rules apply in the order they are specified.
\tool\,home repository contains sample grammars as a cheat sheet. (My personal convention
is to put the commands that process or generate \tool\, commands in `\verb~.tbc~' files, for `the bench command'.
The `\verb~@ command~' can take them as macros to run.)

The first kind of element in (\ref{ex:3el}) starts with space-separated sequence of \textsc{utf-8} text, ending with `\verb~|~'.
Multi-word entries, non-words, parts of words, punctuation and diacritics are possible. (This is provided so that we can avoid single or double-quoted strings at any cost; 
there is no universal programming practice about strings, and the concept is overloaded.)

The material before `\verb~|~' is considered to be the textual proxy of the phonological material of the element.
The next piece is the part of speech for the element, which can be any token (see basic glossary at the back). The material after `\verb~::~' is the category of the element. (This token is the most common convention in monads; it is
called the monadic type constructor.) The part before the single `\verb~:~' is called the syntactic type, which is the domain of s-command. The part after `\verb~:~' is called the predicate-argument structure, which is the domain of l-command. 

Left-associativity is assumed for content, both for  s-command and l-command. Right-associativity is assumed for lambda bindings. 
 Together they constitute a syntax-semantics correspondence for the element. If you don't like such associativity and want to write the structure yourself, please be careful
with the proliferating numbers of parentheses. The system will warn you about mismatches, but the process of getting them right can be painful. In training and exploration with complex elements  it's easy to make judgment errors in parenthesization. With associativity, we can see that 
for example the l-command of first entry in (\ref{ex:3el}) is equivalent to \verb~(like x)y~, in asymmetric tree notation as follows. Its lambda binders
are not part of the command relation.

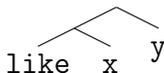
\Tree [[ \verb~like~ \verb~x~ ] \verb~y~ ]
 
 The correspondence is explicit in the order of syntactic slashes and lamba bindings. 
For the first example in\xref{ex:3el}, `\verb~\x~' in the lambda term corresponds to `\verb~/^np~', and `\verb~\y~' to `\verb~\^np[agr=3s]~'. 
The `\verb~\~' in a predicate-argument structure is for `lambda'. Several lambdas can be grouped to write a single `\verb~.~' before the start of a lambda body, as we did in the example.
We could also write them individually, as `\verb~\x.\y.like x y~'.

The syntactic type can be a (i) basic or (ii) complex category. (The term `category' is quite commonly used 
also for the `syntactic type' when no confusion arises.) A basic category is one without a slash, for example `\verb~np[agr=3s]~'.
Here, `\verb~np~' is the basic category and `\verb~[agr=3s]~' is its feature. Features are optional; they can be associated with basic categories only (unlike many other phrase-structure theories). 
Multiple features are comma-separated. An unknown value for a feature is prefixed with `\verb~?~'. For example
`\verb~np[person=1,number=sing,agr=?a]~' means the agreement feature is underspecified. 

Complex categories must
have a slash, e.g. `\verb~s\^np[agr=3s]~' in the example. It is a syntactic function onto `\verb~s~' from `\verb~\^np[agr=3s]~'. The result is always written first, so for example `\verb~s/np[agr=3s]~' would be a syntactic function 
onto `\verb~s~' from `\verb~/^np[agr=3s]~' . Surface directionality would be different. In the first case, the
function would look to the left, in the latter, to the right, in phonological order. 

Modal control
on the directionality is optionally written right after the syntactic slash, such as `\verb~^~' in the first entry. If omitted, it is assumed to be the most liberal, that is
`\verb~.~'. Slash modalities are from \cite{baldridge02,steedmanbaldridge-guide}. \tool\, notation for the modalities
are `\verb~.~' for  `\verb~.~', `\verb|^|' for `$\diamond$',  `\verb|*|' for `$\star$',  and `\verb|+|'  for `$\times$'. They are for surface-syntactic control of  composition in the monad.

Two more tools of \tool\,are worth mentioning; they are not in every categorial grammar's toolbox.  There are double slashes `\verb~\\~' and `\verb~//~' which are similarly backward and forward. They take and yield potential elements of grammar. (Therefore they are unlike Montague's multiple slashes, and they are not necessarily morphological.) 
There is no modal control on them. Basic categories can be singletons, that is, they can stand for their own value only.
For example, \cat{(S\bs NP)/'the\ bucket'} has the singleton `\verb~the bucket~' which stands \emph{as category} for the phonological sequence 
\verb~the bucket~; cf. \cat{NP}, which stands for many expressions that are distributionally NPs. Singletons can be single- or double-quoted as long as they are atomic (i.e. two successive single quotes do not make a double quote).
Singletons are applicative  because they cannot be the range of a complex category, they are domains only ; see \cite{bozs:22-jlli}. 

In summary, elementary items bear categories that are functions of their phonological form.

The second kind of element in (\ref{ex:3el}) is an asymmetric relational rule. It means that
the element bearing the category on the left of `\verb~-->~' at surface form \emph{also} has the category on the right.
The token immediately after `\verb~#~' is the name of the rule, in this case `\verb~np-raise~'. For such rules to make sense, the first lambda binding on the right
must be on the whole predicate-argument structure on the left. This is not checked by the system, we can write any category in principle; but, this is why such rules are not associated with a particular substantive element. Notice that there is no phonological element associated with the rule.

The third kind of element in (\ref{ex:3el}) is a symmetric relational rule. It means that
during analysis either form is eligible for consideration along with its particular category. The analyst apparently
considered them to be grammatically related so that they are not independently listed. (This is one way to capture morphological paradigms but not the only way.) The rule name is right after `\verb~#~'. The  material
on the left and right of `\verb~<-->~' must also contain some phonological material (whitespace-separated words ending with a comma) so that we can reflect the substantive adjustment on the categories. 

There are no elements or rules  in monadic analysis which can alter or delete material in surface structure because that would not be always semantically composable. (It follows that
IA/IP/WP morphology of Hockett must be sprinkled across the three kinds of specification above. Judging from the fact that no morphologically involved language is exclusively IA, IP or WP, this should not be surprising.)

No element type can depend on or produce  phonologically empty elements, because such elements cannot construct categories. The monad's hermetic seal, that every analytic step is atomic, is also consistent with these properties.
 Together they allow us to adhere to the Sch\"onfinkelian idea
of building {hirerarchical structures} from sequential asymmetries alone.%, which is in our case grammatically captured.

\section{Analytics}
Tripartite organization of every synthetic element is reflected in  \tool\,notation:\smallskip
\ex
\verb~phonology :: s-command : l-command~ 
\xe\vskip .5em

It is transmitted unchanged in analytics. Self-distribution of morphology as it sees fit in a grammar
is based on the common understanding that syntax, phonology, morphology and semantics are obviously not isomorphic, but, equally obviously, related, in conveying reference.
%One claim of \mg\,is that they must be homomorphic to analytic structure because of that.

The analytic structure is universal. This means that  these domains are homomorphic to analytic structure, because there is only one such structure.
It is function composition on the semantic side. How the forms (syntax, phonology and morphology) cope with that structure and limit it  is the business of a linguistic theory.

Information transmission in analytics therefore needs a closer look. In building structures out of elements one at a time, the method of matching basic and complex categories of s-command is the following.

If we have the two-category sequence (a) below, we get
(b) by composition, not (c); cf. \verb|f2| passing.

\pex
\a \verb|s[f1=?x1,f2=v2]/s[f1=?x1]  s[f1=v1,f2=?x2]/np[f2=?x2]|
\a \verb|s[f1=v1,f2=v2]/np[f2=?x2]|
\a\verb|s[f1=v1,f2=v2]/np[f2=v2]| 
\xe
This avoids unwanted generation of pseudo-global feature variables, and keeps every step of combination local. Using this property we could in principle compile out all finite feature values for any basic category in a grammar and get rid of its features (but nobody did that,  not even GPSG. Maybe encoder-decoder transformer nets can make a difference here). 
 
Meta-categories such as \cgf{(X\bs X)\fs X} are written in \tool\, as \verb+(@X\@X)/@X+. They are allowed with application only.
Linguistically, only coordination and coordination-like structures seem to need them, which are syntactic islands with \cite{ross67} escape hatches on syntactic identity, that is,
things having the same case, therefore same category:
\ex
\verb~ and | x ::  (@X\*@X)/*@X : \p\q\x. and (p x)(q x)~\\
\xe
This way we can avoid structural unification to do linguistic work; all of it has to be done by universal analytic structure.
(Recall also that structural-reentrant unification is in fact semi-decidable. We simply use term unification.)

For l-command, one important property is that its realization in analysis arises from the evaluation of the surface correspondents in s-command and l-command,
that is, surface structure and predicate-argument structure, therefore we cannot have a complex s-command (i.e. one with a slash) and simplex l-command (i.e.
no lambda abstraction to match the complex s-command). For example, in the following, the `\verb~\np~' has no l-command counterpart (some lambda),
therefore unable to keep the correspondence going:
\ex\label{ex:bad}
*\verb~slept :: s\np : sleep someone~
\xe

However, the inverse, that is, having a complex l-command and simplex s-command, is fine, and common, for example \verb~ man :: n : \x.man x~. This is because the analytics is driven by the syntactic category, that is, by s-command. Not all lambdas have to be syntactic, but all syntactic argument-taking must have a lambda to keep the correspondence going in an analysis. (\ref{ex:bad}) is not an analytically interpretable
synthetic element. If the intention here were to capture topic-dropped `sleep,' for example \emph{slept all day, what else could I do?}, it might be 
say \verb~slept :: s : sleep topic~.
\section{Exploration}
\paragraph{1.}There are some aids in \tool\,to explore monadic analytic structures. We can explore
what kind of syntactic case and similar structural functions arise given a grammar, from its verbs and verb-like elements. These functions can be conceived as asymmetric relational rules, showing
the understanding that if we have one category for an argument in an expression we also have other categories for the argument, as a sign of
mastery of argument-taking in various surface expressions. 

The `\verb~c command~' of \tool\,generates these functions from a grammar (loaded by the `\verb~g command~'), taking a list
of parts of speech as input from which to generate the second-order functions. (Presumably, these are the parts of speech
of the verbs and verb-like elements.)
These extra functions are saved
in a separate textual grammar file with extension `\verb~.sc.arules~'. The name
designates that all of them are asymmetric relational rules. 

These rules can be merged manually with the
grammar from which they are derived. The reason for not automating the merge is because you may want 
to play around with them before incorporating them  into an analysis. It is a textual file, therefore editable
just like any grammar text file.
Many surface structures follow from incorporating them into a grammar, which you can check with the `\verb~a command~' (for `analyze').

After the merge, analyses will reveal many more surface configurations as a consequence of synthetic case  in the language under investigation.
This way of looking at case renders generalizations about case, for example lexical, inherent and structural case \citep{Woolford:06},
as bottom-up typological universals arising from a range of class restrictions on the verb, from most specific to least.%\mg\,has more to say about these aspects. 

\paragraph{2.}We can preview the syntactic skeleton of a grammar in \tool. `\verb~k command~' goes over all the elements of the currently loaded
grammar, and reports how many distinct syntactic categories we have in it.

 For brevity it does not report features of a basic category. 
This much is certain though: if two  categories reported by this command look the same but reported separately, it would mean
that they are distinct in their features, to the extent that they would not match in analysis either.
For example, \verb~s\np[agr=3s]~ and \verb~s\np[agr=1s]~ are distinct, although their featureless version looks common: \verb~s\np~.

One aspect of the `\verb~k command~' can be used to find out patterns in the grammar's categories, and to find out why some categories are distinct eventhough they might look identical as reported by the `\verb~k command~': for every distinct
syntactic category reported, it enumerates the list of elements that bears it. These forms are presumably
indicators of common syntactic, phonological and/or morphological properties, for example agreement classes and bound versus free elements.

\tool\,is a tool for a very bottom-up view of grammar. Naturally, more than N, V, A, P distributions will be reported. (In fact, \tool\,has no
built-in universal categories or features. If you don't use these four categories at all, it is fine.) To see the basic category inventory, in addition
to the inverted list display of the `\verb~k command~', take a look at the output of `\verb~! command~', which reports the basic categories in the grammar and
the list of all features in them, and  attested values.

\paragraph{3.} There is no morphophonological analysis built in to \tool. Morphological boundedness of a synthetic form  can be designated with `\verb~+~' in \verb~a~ and \verb~r commands~.
It goes as far allowing 
in analysis/ranking entering surface tokens such as MWEs either as `\verb~dismiss ed'~ or `\verb~dismiss+ed~', if you have
`\verb~ed~' as an entry in the grammar. (`\verb~+~' is a special operator for this, i.e. the entry
is not assumed to be `\verb~+ed~'.) Keep also in mind that \tool\,has no notion 
of `morpheme'---you must model that if you want---but it has a notion of morphological structure: the category building on phonology to construct reference, Montague-style.

The current `+' notation is just for some convenience until somebody comes up with
a categorial theory of morphophonology for example along the lines of \cite{Schm:81,hoeksemajanda88}. Till then, we won't get `\verb~insured~' from `\verb~insure+ed~', or Turkish harmony distinction`\verb~avlar~' (game-PLU) 
and `\verb~evler~' (house-PLU) from e.g. `\verb~av+ler~' and `\verb~ev+ler~'.

\paragraph{4.}We can turn the textual form of \tool\,grammar into a set of data points each associated with a parameter. 
These parameters 
are parameters in the modeling sense, characterizing a data point. To do this, \tool\,turns a textual grammar into a `source form', which is a \textsc{Common Lisp} data structure, basically Lisp source code.
Such files are automatically generated and carry the file extension of `\verb~.src~'. 

This is an inspectable file, but it is much easier to
inspect when it is turned into a textual grammar in \tool\,notation, which the `\verb~z command~'  does.\footnote{The `\verb~z command~' can also transform legacy \verb~`.ded', `.ind~' and `\verb~.ccg.lisp~' grammars of CCGlab to \tool\,  grammar format. Please
change the top line of such grammars to `\verb~(defparameter *current-grammar*~' and put the result
in \verb~/var/tmp/thebench~ directory before running the command. The result will be put back in your working directory as a text file in the monadic format.}
The resulting textual file will be much like the original textual grammar, without comments, and with keys and initial value of the parameter added on the right edge of every entry, in the form
of for example `\verb~<314, 1.0>~' where 314 is the key and 1.0 is the parameter value. The keys are unique to each entry, therefore
having multiple categories for the same form is possible; they would be distinct entries even if they are identical in text.

These values are not probabilities; they are weights.
The ranker (the `\verb~r command~') turns them into probabilities to arrive at most likely analyses, given the weights. The trainer adjusts these weights
depending on supervision data.
Therefore it is important to know that
entries \emph{with the same phonological form} compete with their weights in making themselves enter an analysis. (In other words,
there is no built-in determinism, mapping each phonological form to only one interpretation.)

The system makes the initial  assignment automatically  
to ensure the uniqueness of the keys and completeness of parameter assignment.\footnote{If you use a specific key for a particular entry in your grammar text, say for easier
 tracking of a particular item during model training, the system respects your key choices and parameter initialization. I recommend not using this feature extensively
to avoid key clashes. Uniqueness of personally assigned keys is not checked. If you do use the special key assignment feature,
make sure that there is no keyed element after any unkeyed element, that is, pile unkeyed elements at the end of the file. This way your assigned keys will be seen before new key assignment begins.}
For example, when you inspect the regenerated textual grammar containing (\ref{ex:3el}), you may see something like:

\ex\label{ex:3elkeys}
\footnotesize
\begin{tabular}[t]{l}
 \verb~likes  | v :: (s\^np[agr=3s])/^np : \x\y.like x y <120, 1.0>~\\
\verb~#np-raise np[agr=?x] : lf  --> s/(s\np[agr=?x]) : \lf\p. p lf  <34, 1.0>~\\
\verb~runs | tense :: s[t=pres,agr=3s]\np:\x.pres run x  <2, 1.0>~\\
\verb~ran | tense :: s[t=past]\np:\x.past run x <76, 1.0>~
\end{tabular}
\xe
This file is just as editable and processable as the original textual file containing the grammar. However, notice the difference from (\ref{ex:3el}).
The symmetric relational rules are compiled into separate entries in (\ref{ex:3elkeys}), and their link is preserved by using the name of the rule
as  `part of speech' in both entries, in this case \verb~tense~. 

The point of creating different data points is that, when trained, these elements will participate in different analyses therefore need  parameter values of their own. Using the preserved link (same tag) we can explore reconstruction of the paradigm intended by `\verb~<-->~'. The possibility of using the same rule name to capture a paradigm facilitates this exploration.

There is one more change in the reconstructed text file: every entry's capitalization is normalized to lower case, except the phonological material, whose ``orthographic case'' is preserved. This
is one lame attempt to indicate that \textsc{Common Lisp}'s default capitalization of values without us asking for it
would not make any difference to an analysis.\footnote{I could make it case-sensitive, to preserve both analyst's and Lisp's cases,
but this would be quite error prone: Is \verb~AdvP~ same as \verb~advp~? I would think so.}

\paragraph{5.}Taking the text source of a developing grammatical analysis and turning to modeling with it is the idea
behind the textual reconversion of a grammar source. The reconstructed source has the symmetric rules
turned into separate entries with same `part of speech': the rule's name.  They will have their own parameter
values. This is probably the last step
after grammar development, when it is time to move on to grammar training to adjust belief in elements, using parameters. This is why
the command that does this is called the `\verb~z command~'.

Training a grammar with  data updates the parameters, which can then be used for ranking the analysis, that is, for choosing from the possible analyses the most likely one after training.
The method used is sequence learning of \cite{zettlemoyercollins05}. It is basically a gradient ascent method. It places the whole trained grammar 
in probabilistic sample space of possible grammars.\footnote{Linguistics alert: These are not probabilistic grammars, things in which a category would be uncertain. They are grammars with clear-cut categories which are situated by model selection in the probabilistic 
space of possible grammars. The idea is not too far from
the continuity hypothesis in language acquisition by which
a child moves from one possible grammar to another, apparently
nondeterministically; see \cite{Crai:98}.}

Supervision pairs are correspondences of phonological forms and their correct predicate-argument structures,
sometimes called in computational work `gold annotation'. However, the term would be misleading, because there is no annotation or labeled data. 

The term `supervision' needs clarification in the context of \tool. It means that we \emph{presume} that that data is known to hold some correspondence
of expressions and predicate-argument structures, much like \cite{Brow:73}, \cite{brow:98} and \cite{Toma:92} did when they started analyzing child data---it is indeed an adult assumption coming from psycholinguistic analysis.
The predicate-argument structure is a specification of the presumed meaning of the phonological item, written in a text file one line per entry, for example:
\ex\label{ex:sp}\small\begin{tabular}[t]{l}
\verb~Mary persuaded Harry to study : persuade (study harry) harry mary~\\
\verb~Mary promised Harry to study : promise (study mary) harry mary~\\
\verb~Mary expected Harry to study : expect (study harry) mary~\\
\end{tabular}
\xe

Here, the material before the colon is the proxy phonological form (therefore ``orthographic case''-sensitive). Multi-word expressions (MWEs) must be enclosed within `'\verb~|~' if they are considered to be referentially atomic, e.g. \verb~|kicked the bucket|~. (After all, this is supervision in the sense above, so we assume its reference is
fixed, we just need to find out which part of the predicate-argument structure corresponds to it, given a grammar with that MWE.)

The material after the colon is the
presumed predicate-argument structure of the whole expression. It has the same format as in grammar specification.
The `\verb~t command~' (for `train') updates the parameters of a grammar-turned-into-a-model by this method.
We can pick from the candidate models by looking at its output (known as model selection), which is a collection of plain text grammars (the number of  candidates is specified by you), with parameters for every entry. We can then load the chosen grammar, and rank analyses with it using the `\verb~r command~'.

\keep{The input to training is as many as number of entries in a file with lines like those in\xref{ex:sp}.} We suggest minimal (required) parenthesization to avoid
near misses in training; \tool\,will internally binarize the predicate-argument structure anyway.
\keep{Exact repetition of an entire line constitutes a separate supervision entry.
This is useful---in fact required---in analyzing child-directed speech, where repetition is very common.}\footnote{Assigning different
semantics to every repetition of an identical expression is the holy grail of child research. We have no theories for that as far as I know.}

\paragraph{6.}The tools used for training  a grammar on data such as\xref{ex:sp} are triggered from \tool\,command line, but they work offline, using the same processor code. These training sessions can last quite long, sometimes hours, sometimes days or weeks depending on grammar size, number of experiments attempted and amount of supervision data.
(That's because of the computer speed.)
We don't have to stay in the originating session to keep the experimental runs going. They operate independently, using the \verb~nohup~ facility.\footnote{\verb~nohup~ is for `no hang up', which is one more reason why a \verb~linux~ system or subsystem is
indispensable for tools such as this. One caution: Do \emph{not} exit your session with control-D, which would kill the subprocesses in the linuxsphere. Just let it die or hang, or close the window normally, e.g. by clicking the red button in the red-yellow-green window menu.}  

During these runs, all information must be available offline. To do this, we  prepare an `experiment file'. It has a strict format, for example:

\ex\label{ex:expfile}
\begin{tabular}[t]{l}
\verb~7000 4000 xp 1.2 1.0 nfp nfparse-off~\\
\verb~4000 2000 10 0.5 1.0 bon beam-on~\\
\verb~4000 2000 10 0.5 1.0 boff~
\end{tabular}
\xe

This specification will fetch three processors if it can because three lines of experiments are requested, one using 7GB of RAM in which 4GB is heap (dynamic space needed for internal structures such as hash tables, etc.), the others with 4GB RAM and 2GB of heap. 
As a guideline, a larger heap helps to run analyses of longer expressions, which
usually arises when case functions are added to a grammar.

These parameters are useful to run experiments on machines with different powers  and architecture (number of cores, amount of memory, size of heap) without
changing the basic setup.

The `\verb~t command~' specifies which grammar and supervision will take place in the experiments. (Therefore
every experiment in one file runs on the same grammar-training data.)
The supervision file must be in the format of (\ref{ex:sp}). The rest are training parameters: \verb+xp+ means
use of the extrapolator, which uses a predetermined number of iterations for the gradient ascent in implementing the \cite{caba:76} method.  

The second experiment does not extrapolate; it iterates 10 times. Every iteration updates the gradient incrementally.
Then comes the learning rate (1.2 in the first experiment), which is the distance the gradient travels in one step (`the jump distance'),
the learning rate rate (1.0 in the first experiment), which is how later iterations affect the jump distance, the
prefix of the log file e.g. \verb~nfp~ (the actual name of the log file is prefixed with that, adding training parameters as suffixes of the name for easier identification of experiments that will be saved in the end), and the function to call before training starts, which is optional, as in the third experiment. In the first experiment it is \verb+nfparse-off+, which turns normal form parsing off, which is a feature in the processor, implementing the algorithm of \cite{eisner96} for reducing  compositions after they come out of the monad. This affects the number of analyses generated and reported. 

The full list of such functions
is given at the back, which are callable from the command interface as well  the `\verb~l command~'. (For processor functions
with  more than one arguments, calling is a bit more complex; see \verb~cl4py~ Python library documentation.)
In the second experiment, it is \verb~beam-on~. This function focuses the gradient on items with largest weight changes during iterations, avoiding consideration of elements that change very little (no change is effectively handled by default). The results may be less precise because of this but it reduces the search space for the gradient. It is sometimes a lifesaver in eliminating
excessive runtimes or out-of-memory errors. 

\paragraph{7.}Studying referentiality of all kinds along with configurationality in one package seems to be one way to avoid the purportedly dichotomic world of verbs-first and nouns-first approaches in language acquisition. \tool\,is designed primarily to rise above these assumptions in modeling.

By combining explorations \textbf{4--5--6} it is possible to have another look at event-and-participant attempts to explain language acquisition, for example \cite{Brow:73,Toma:92,brow:98,MacW:00,Aben:15a}, to compare with participant-centric approaches such as \cite{gentner82}.

In such an undertaking, the observables would be the phonological form on the left, much like in\xref{ex:sp}, and the analyst's conclusion about what they would mean
would be the predicate-argument structure on the right. There would be no labeled or hidden variables such as syntactic labels or dependency types.
The mental grammar which is presumed to arise in the mind of the child would be proxied as a grammar much like in the earlier sections, which would
be trained on the data such as above for model selection.

\section{Work cycles}\label{sec:work}
Three work cycles are reported here from personal experience, for whatever its worth. 

The first one is for the development of an analysis by checking
its aspects with respect to theoretical assumptions. This would be the simple  cycle in (\ref{ex:3cycles}a), with `\verb~k~' and `\verb~!~' commands sprinkled in between (not shown here). 
\pex\label{ex:3cycles}
\a \verb~edit - g command - a command - , command~
\a \verb~edit - g command - c command - edit/merge - a command - , command~
\a \verb~edit - t command - z command - g command - r command - # command~
\xe
It would load a grammar (~\verb~g~), analyze an expression with it (\verb~a~), and display results (\verb~,~).

The second cycle (b) would be for studying a grammatical analysis in its full implications for surface structure when the class of verbs is large enough. 
`Merge' here presumably  merges case functions with the grammar. When we generate case functions using `\verb~c command~', the currently
loaded grammar have access to them so we can do  analyses with them in the current session. However, they are not added
to grammar text file, so the grammar is left intact when the session is over. Merge is left up to you. The case functions are saved in a file to facilitate that.

The third one, (c),  is for training a grammar to see how it affects ranking of analysis. %This would be the cycle in (c). 
The `\verb~z command~' here presumes that one trained grammar is selected for ranking, so the outputs of the `\verb~t command~' have 
presumably gone through some kind of model selection after training, that is, choosing one of the grammars with updated weights, either manually or by a method of model selection. 

The `\verb~t command~' in (c) will fork as many processes as the number of experiments requested; NB. discussion around\xref{ex:expfile}. It's best to use it offline. For that,
put it in commands file and use the `\verb~@ command~' for batch processing. Waiting online can be painful depending on the number of experiments and data sizes.
Just don't do control-D on \tool\, interface if you use it online; it will kill the subprocesses. Let the terminal hang. Don't close the lid on a laptop; that would suspend all work. (Speaking of laptops, which are more and more designed toward efficient handling of video, audio and graphics using
specific processors for them, it is best to use a more general high-speed multicore/multiprocessor system to do the training. Many experiments
that ground a laptop to a halt worked effortlessly when I ran them on a powerful desktop.)

If you are training a grammar on a  large supervision dataset, and want to find out how top-ranked analyses fare with gold pairings of expressions and meanings, one option 
of `\verb~# command~' combined with the `\verb~> command~' can be helpful. Before ranking, turn on logging with the `\verb~> command~'. Then for `\verb~# command~' in a batch command file use the `\verb~bare~' option to eliminate verbosity. Minimal amount of
extra text will be saved in the \emph{processor's log file}, which you can easily eliminate. Don't forget to turn off logging at the end, using the `\verb~< command~'. 

One advice about the `\verb~/ command~'. This command clears the ~\verb~/var/tmp/thebench~ directory where internal results of \tool\, are kept.
Do not use this command if you started an experiment, either online from the interface or in batch mode (see glossary). These commands create
files that are needed later on; see the end of \S\ref{sec:org} for file organization in \tool.

\section{To do}
I have a personal to-do list for collaborative work. I mention here the technical ones which require programming. For typology work, just drop me an email please.

\paragraph{1. Implementing the `dashed boxes' of \tool.} `Analysis' as understood in \tool\,is a wholistic process, where every part-process is interpreted in the light of the whole in check all the time, at every step. Analyses can be displayed 
in the following form.
\ex\label{ex:rc-ncn0}
\begingl
\gla {yaac-waas-it-\textglotstop i\v{s}} {\v{c}akup-\textglotstop i}
yaq-it-ii {\textglotstop uut'yaap} suu\textsubdot{h}aa//
\glb walk-outside\xg{-past-3sg} man\xg{-det} \xg{-rel-past-3sg.rlt} bring salmon//
\glft `The man who brought salmon left.'\trailingcitation{Nuu-Chah-Nulth; \citealt{wojd:00}:274}//
\endgl\\
{\pagodasize
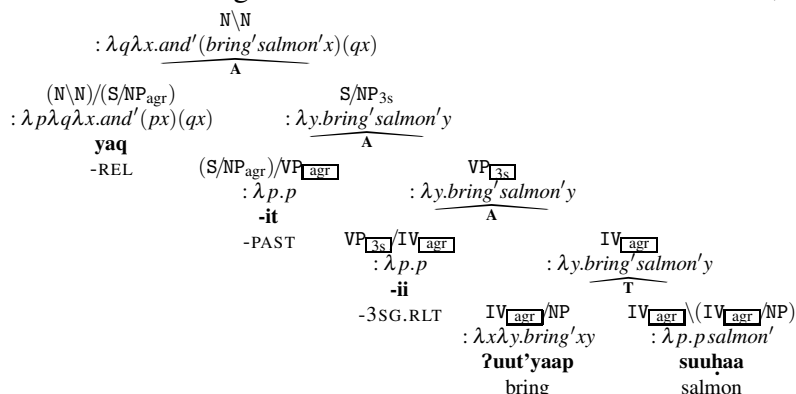
\begin{smex} % S/NP3s
\begin{smex} % N\N
\smres{\cat{N\bs N}\\ \lf{\lambda q\lambda x.\so{and}(\so{bring}\so{salmon}x)(qx)}}{\comba}{4em}{4em}\\
\smleaf{\cat{(N\bs N)\fs(S\fs\cgs{NP}{agr})} \\ 
             \lf{\lambda p\lambda q\lambda x.\so{and}(px)(qx)}\\ \phon{yaq} \\ \xg{-rel}} &
\hspace*{-3em}\begin{smex}
\smres{\cat{S\fs\cgs{NP}{3s}}\\  \lf{\lambda y.\so{bring}\so{salmon}y}}{\comba}{3em}{4em}\\
\smleaf{\cat{(S\fs\cgs{NP}{agr})\fs\cgs{VP}{\bfeat{agr}}}\\ \lf{\lambda p.p}\\ \phon{-it}\\ \xg{-past}} &
\hspace*{-2em}\begin{smex} % VPagr
  \smres{\cat{\cgs{VP}{\bfeats{3s}}}\\  \lf{\lambda y.\so{bring}\so{salmon}y}}{\comba}{3em}{3em}\\
  \smleaf{\cat{\cgs{VP}{\bfeats{3s}}\fs\cgs{IV}{\bfeat{agr}}}\\ \lf{\lambda p.p}\\ \phon{-ii}\\ \xg{-3sg.rlt}} &
      \hspace*{-2em}\begin{smex} % IVagr
        \smresl{\cgs{IV}{\bfeat{agr}}\\ \lf{\lambda y.\so{bring}\so{salmon}y}}{\combt}{2em}\\
        \smleaf{\cat{\cgs{IV}{\bfeat{agr}}\fs NP}\\ 
               \lf{\lambda x\lambda y.\so{bring}xy}\\ \phon{\textglotstop uut'yaap}\\ bring} &
        \smleaf{\cat{\cgs{IV}{\bfeat{agr}}\bs(\cgs{IV}{\bfeat{agr}}\fs NP)}\\ 
                  \lf{\lambda p.p\,\so{salmon}}\\ \phon{suu\textsubdot{h}aa}\\ salmon}
\end{smex}
\end{smex}
\end{smex}
\end{smex}
\end{smex}
}
\xe
However, the original conception of categorial grammar was not just a better and simpler alternative to phrase-structure grammar, with or without movement. (I suppose this rather unfortunate overinterpretation emanated from \citealt{BarH:60}.)  
Every step of analysis is supposed to keep the whole in check, with parts showing \emph{subprocesses} themselves.
This was the connection to Gestalt Psychology, see for example \cite{wert:24}.

So what we really want to implement in \tool\,is showing how the whole is kept in check at every step. The dashed boxes
in the following analysis of the same example show how it  can be done. It has not been implemented in \tool\,yet.

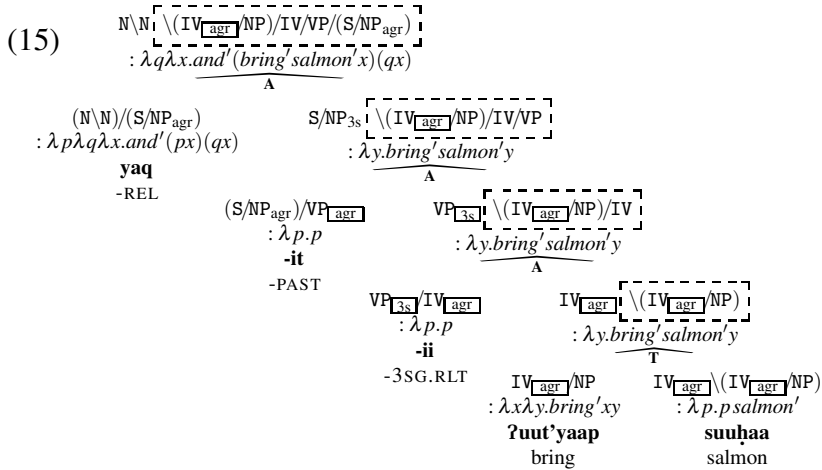
\ex
{\spagodasize
\hspace*{-3em}\begin{smex} % S/NP3s
\begin{smex} % N\N
\smres{\cat{N\bs N}\,\dbox{\cat{\bs(\cgs{IV}{\bfeat{agr}}\fs NP)\fs IV\fs VP\fs(S\fs\cgs{NP}{agr})}}\\ \lf{\lambda q\lambda x.\so{and}(\so{bring}\so{salmon}x)(qx)}}{\comba}{4em}{4em}\\
\smleaf{\cat{(N\bs N)\fs(S\fs\cgs{NP}{agr})} \\ 
             \lf{\lambda p\lambda q\lambda x.\so{and}(px)(qx)}\\ \phon{yaq} \\ \xg{-rel}} &
\hspace*{-3em}\begin{smex}
\smres{\cat{S\fs\cgs{NP}{3s}}\,\dbox{\cat{\bs(\cgs{IV}{\bfeat{agr}}\fs NP)\fs IV\fs VP}}\\  \lf{\lambda y.\so{bring}\so{salmon}y}}{\comba}{3em}{4em}\\
\smleaf{\cat{(S\fs\cgs{NP}{agr})\fs\cgs{VP}{\bfeat{agr}}}\\ \lf{\lambda p.p}\\ \phon{-it}\\ \xg{-past}} &
\hspace*{-2em}\begin{smex} % VPagr
  \smres{\cat{\cgs{VP}{\bfeats{3s}}}\,\dbox{\cat{\bs(\cgs{IV}{\bfeat{agr}}\fs NP)\fs IV}}\\  \lf{\lambda y.\so{bring}\so{salmon}y}}{\comba}{3em}{3em}\\
  \smleaf{\cat{\cgs{VP}{\bfeats{3s}}\fs\cgs{IV}{\bfeat{agr}}}\\ \lf{\lambda p.p}\\ \phon{-ii}\\ \xg{-3sg.rlt}} &
      \hspace*{-2em}\begin{smex} % IVagr
        \smresl{\cgs{IV}{\bfeat{agr}}\,\dbox{\cat{\bs(\cgs{IV}{\bfeat{agr}}\fs NP)}}\\ \lf{\lambda y.\so{bring}\so{salmon}y}}{\combt}{2em}\\
        \smleaf{\cat{\cgs{IV}{\bfeat{agr}}\fs NP}\\ 
               \lf{\lambda x\lambda y.\so{bring}xy}\\ \phon{\textglotstop uut'yaap}\\ bring} &
        \smleaf{\cat{\cgs{IV}{\bfeat{agr}}\bs(\cgs{IV}{\bfeat{agr}}\fs NP)}\\ 
                  \lf{\lambda p.p\,\so{salmon}}\\ \phon{suu\textsubdot{h}aa}\\ salmon}
\end{smex}
\end{smex}
\end{smex}
\end{smex}
\end{smex}
}
\xe
The overall result on top in its full form does not mean that for example \cat{(S\fs\cgs{NP}{agr})} is followed in sequence by \cat{VP}, then \cat{IV}, and so on. It shows
the structural unfolding of the entire expression, that is, the part-processes.  

Such structures would be necessary but not sufficient to situate reference in context.

\paragraph{2. Production/generation.} Categorial production can be seen as expressing an intended complex reference by choosing elements from grammar that would serve that reference. It would be more than expression of thought. It is more like environment control by language.

One way to start studying such processes is to see how far compositionality in grammar can be put to work on this task. To do this,
it would be nice to have a ``structural chunker,'' one that would take the triplet \verb~<word, category, size>~ and produce an expression, where
\verb~word~ is the chosen phonological left edge of the chunk, \verb~category~ is the syntactic category of the overall result intended, and \verb~size~ is the upper limit on the chunk in terms of number of elements in it. The upper limit would be needed
to make the problem decidable because the category can be endotypic, for example \cat{S\bs S}, therefore the process is potentially recursive.

Generation is usually understood that way in language technology with phrase-structure grammars, because they  tend to take care of distributional adequacy first and worry about reference later. In categorial grammar however, it is half the story, because reference itself would reveal item choice in the first place. %So it is number 2 in my to-do list.

\paragraph{3. Encapsulating the monad.} A true monad is a hermetic seal. We cannot peek inside to see 
intermediate results. In the current  state of \tool, we can. That's because it is easier to debug.

However, turning the current state 
of the implementation to a true monad requires more than employing a monad library in Python or Common Lisp.
The categories that are input to the monad would have to be lifted in syntax and semantics. 

%\paragraph{4. Cheat sheet.} I have made up lots of conventions to make \tool\,usable by others, for example, naming conventions, capitalization, MWEs, etc. I provide a glossary of the ones I could think of at the back.
%It would be nice to have a concise list of these conventions, with pointers for things to look out for. Program designers are not good at coming
%up with  cheat sheets for their own mess (last time I did that I got it wrong, even worse, misleading). I ask for help from experienced users.
%In the good old days it used to be students who use the `lab tool' of the course for grade doing this, but nowadays this is considered slave labor.

\section*{Acknowledgments}

I thank Python, Lisp, \texttt{StackOverflow} and \texttt{StackExchange} communities for answering my questions before I ask them. 
The predicate-argument structure evaluator in the processor 
is based on Alessandro Cimatti's abstract data structure  implementation of lambda calculus in Lisp, which allows us to see the internal make-up of
l-command. 
David Beazley's \texttt{sly} python library made \tool\,interface description a breeze.  Marco Heisig's \texttt{cl4py} python
library made Python-Lisp communication easy, which allowed me to recycle some Lisp code. Luke Zettlemoyer explained
to me his sequence learner in detail so that I can implement it on my own. Discussions with Chris Stone clarified some of the tasks
in to-do list.
Thanks to all five colleagues. I  blame good weather, hapless geography and my cats for the remaining errors.

\newpage

\setlength{\parskip}{-4pt}

\section*{Basic glossary of \tool} 
\paragraph{analysis} Combining synthetic elements of grammar in a given expression by composition.
\paragraph{auxiliary file} A file internally generated by \tool\,for the processor, and saved in \verb~/var/tmp/thebench~. 
Examples are grammar source as Lisp code
and processor-friendly supervision files.
\paragraph{batch mode} Use of the `\verb~@ command~'. You will see \tool\,prompt doubled in the log file.
The commands in command files can themselves be the `\verb~@ command~; useful for trying many models at batch mode.
\paragraph{editable file} A grammar file that you create and save as text, or the grammar files that \tool\,generates
by `\verb~c command~' and `\verb~z command~'.
\paragraph{element} A (synthetic)  element of grammar; one of (\ref{ex:3el}).
\paragraph{grammar, case functions file} A file automatically generated by \tool\,when you run the `\verb~c command~'. File extension is `\verb~.sc.arules~'. Such files are editable, therefore saved in your working directory. 
Merging these files with grammar text is up to you. The `\verb~c command~' adds them to currently loaded grammar only.
\paragraph{grammar, text file} A textual file containing a grammar with entries in the form of (\ref{ex:3el}).
File extension is optional, and it is up to you. Resides in your working directory.
\paragraph{grammar, re-text file} A textual file generated by the system from a `\verb~.src~' file.
The result file is identical in structure to the edited grammar which generated the `\verb~.src~' file. The system adds a default parameter value and a unique key to identify every entry. Such files are saved in your working directory after fetching the `\verb~.src~' file from \verb~/var/tmp/thebench~, indicating that they are editable. The `\verb~z command~' does this. 
\paragraph{grammar, source code file} A textual grammar which is transformed to the  processor notation. It is actually Lisp source code. Its extension is 
`\verb~.src~' if generated by the system. \tool\,saves them in the folder \verb~/var/tmp/thebench~ by default.
\paragraph{identifier, token} An atomic element of \textsc{ascii} symbols with at least one alphabetical symbol, not beginning with one of \{\verb~+, *,^, .~\}, which are the slash modalities. It may contain and begin with: a letter, number, tilde, dash or underscore.
Examples are basic categories, feature names and values, rule names, parts of speech, predicate basic terms (names of
predicates and arguments), lambda variable names.
\paragraph{identifier, predicate modality} An identifier that consists of `\verb~_~' only. 
It is a simple convention from \cite{bozs:22-jlli}, nothing universal, to signify that the stuff before
it is the predicate and the stuff right after it is a modality of the predicate rather than an argument. Used in l-command only.
\paragraph{identifier, variable} An identifier that begins with `\verb~?~'. Used in feature values of basic syntactic
categories to represent underspecification.
\paragraph{!identifier, !token} An identifier/token in a lambda term prefixed with `\verb~!~'. The Lisp processor converts
such identifiers to a double-quoted string. For example, \verb~!_name~ becomes \verb~"_name"~ in the processor. It is kept for legacy.
\paragraph{intermediate representation} Sometimes it is difficult to tell whether a textual element
in grammar text has been converted to proper structural representation. Using the `\verb~i command~' you can
check the internal structure of all entries in a grammar in the form of Python code. The resulting file is  editable (e.g. if you want to explore its dictionary data structure in \verb~python~), therefore saved in your working
directory unless specified otherwise. You can do the same check for one element only, without adding it
to grammar. Use the `\verb~- command~' for that.
\paragraph{item} Any space-separated sequence of \textsc{utf-8} text, for example the phonological material. In  representation and processing, it is case-sensitive.
\paragraph{MWE} Multiword expression. Something referentially atomic in a surface expression, marked as \verb~|..|~, for example \verb~|the bucket|~. Whitespacing is not important. In a grammar entry, multiple words before the part of speech is by definition an MWE.
\emph{In a surface expression}  they must be within \verb~|..|~ to match that element in analysis.
\paragraph{modality, syntactic} One of \{\verb|^|, \verb|*|, \verb|+|, \verb~.~\}   after a syntactic slash as a further surface constraint.
Assumed to be  `\verb~.~' if omitted. Controls the amount (none `\verb~*~' to all `\verb~.~') and degree (harmony `\verb~^~' disharmony `\verb~+~') of composition. 
\paragraph{part of speech tag} An identifier which can be put to various use, e.g. morphological identification, grammatical organization. Such identifiers can be any sequence of ASCII non-space tokens.
No universal set is assumed. For example, the `\verb~c command~' uses these tags to derive case functions, because verbs cannot be discerned from their syntactic category alone. Take for example the category \cat{(S\bs NP)\bs VP}:
Is this a verb or an adverb?
Somebody  has to just designate
them as verb or verb-like, to be targeted by the `\verb~c command~'. 
\paragraph{ranking} Looking at most likely analysis of an expression given a trained grammar.
\paragraph{relational rule, asymmetric} A rule which maps a surface category to another surface category, to indicate
that the surface form bearing the first one also bears the second one. In analysis, they apply in the order specified in grammar.
%In specification, the first one is mapped to the second.
\paragraph{relational rule, symmetric} A rule which maps two surface forms  to each other if they are
categorially related. In analysis, they are treated as independent elements; whichever takes part in surface
structure will be used with its own category. Their order is not important, either in rule specification or in analysis.
\paragraph{singleton} A basic category in single or double quotes, in basic and complex categories. For example
\verb~(s\np)/"the bucket"~ treats \verb~"the bucket"~ as a category that can only take on one value,  the surface string itself, as in
\emph{kicked the bucket} kind of idioms, so that a much narrower reference compared to the more general category \cat{NP} can be discerned; see \cite{bozs:22-jlli}. %Currently, \tool's analytics treats such categories as matching
%similarly referentially constrained MWEs in a surface expression, e.g. \verb~kicked |the bucket|~, rather than \verb~kicked the bucket~.
%Presumably \verb~|the bucket|~ covertly refers to a euphemism, whereas the expression \verb~the bucket~ has overt reference.

\paragraph{special category} Any basic category prefixed with `\verb~@~'. They are application-only, to avoid nested term unification.
Useful for coordinands and clitics, i.e. things that can take one complex category no matter what in one fell swoop.

\paragraph{supervision, text file} A textual file containing entries in the form of (\ref{ex:sp}).
File extension is optional; it is up to you. Resides in your working directory.

\paragraph{supervision, source code file} A supervision file generated by the system from your  supervision file text.
It has the extension `\verb~.sup~'. It is in the Lisp format that the processor uses. Such files are saved in the folder \verb~/var/tmp/thebench~.

\paragraph{synthesis} Putting together of an element of grammar. Although an element has components
such as the phonological form, part of speech, the syntactic type and the predicate-argument structure, as far as analytics is concerned, this is a hermetic seal, and only the monad can manipulate the components.
\begin{table}[b]
\caption{\small The command interface of \tool. There is also the phantom command called `\texttt{pass}', which is useful for commenting on the
tasks in the `\texttt{@ command}'s batch processing files. For example, `\texttt{pass the next command generates case functions}' does nothing
but echoes itself. Any command in the interface can be put in batch files as long as it can be processed without online input.}
\scriptsize
\begin{verbatim}
______\ Letter commands are processor commands; symbol commands are for display or setup
_______\______ Dots are referred in sequence; .? means optional; .* means space-separated items
 a .*   | analyzes . in the current grammar; MWEs must be enclosed in |, e.g. |the bucket|
 c .*   | case functions generated for current grammar from elements with POSs .
 e .*   | evaluates the python expression . at your own risk (be careful with deletes)
 g .    | grammar text .  checked and its source made current (.src file goes to /var/tmp/thebench/)
 i .    | intermediate representation of current grammar saved (file . goes to /var/tmp/thebench/)
 k      | reports categorial skeleton of the current grammar---its distinct syntactic categories
 l ..?  | Lisp function . is called with args ., which takes them as strings
 o .*   | OS/shell command . is run at your own risk (be careful with deletes)
 r .*   | ranks . in the current grammar; MWEs must be enclosed in |, e.g. |the bucket|
 t ...  | trains grammar in file . on data in file . using training parameters in file .
 z .    | source . located in /var/tmp/thebench/ and saved as editable grammar locally (.txt)
 @ ..   | does (nested) commands in file . (1 command/line, 1 line/command); forces output to .log
 , .*?  | displays analyses for solutions numbered ., all if none provided
 # .?   | displays ranked analyses; outputs only [string likeliest-solution] pair if . is 'bare'
 = .*   | displays analyses onto basic cats in .
 ! .?   | shows basic cats and features of current grammar (optionally saves to file .log)
 $ .*   | shows the elements with parts of speech in .
 - .    | shows (without adding) the intermediate representation of element .
 + .    | processor adds Lisp code in file .
 > ..?  | Logs processor output to file .log; if second . is 'force' overwrites if exists
 <      | Logging turned off
 /      | Clears the /var/tmp/thebench/ directory
 ?      | displays help
_______/ Use UP and DOWN keys for command recall from use history
\end{verbatim}
\end{table}

\begin{table}
\caption{Processor functions accessible from experiment files and
the command line.}\medskip
{\footnotesize
\begin{tabular}{ll}

\verb|beam-on| & {Turns the beam on. For training.}\\

\verb|beam-off| & {Turns the beam off (default). For training.}\\

\verb|beam-value| & {Shows current properties of the beam processor (status and exponent)}. For display.\\

\verb|lambda-on| & {Turns on display of lambda terms at every step of analysis (default). For display.}\\

\verb|lambda-off| & {Turns it off. Final results are always shown. For display.}\\

\verb~monad-all~ & Processor set to use all monad rules (default). For analysis and training.\\

\verb~monad-montague~ & Processor set to use only application in composition (\comba\, and \combt).\\
&  For analysis and training.\\

\verb|nfparse-on| & {Turns normal-form parsing on (default). For display.}\\

\verb|nfparse-off| & {Turns normal-form parsing off. For analysis and training.}\\

\verb|onoff| & {List of on/off switches that control analysis.}\\

{\verb|oov-on|} & {Turns on  out-of-vocabulary treatment. Two dummy entries assumed for OOV items:}\\
& One with the category  \verb~@X\@X~ and the other with \verb~@X/@X~. For analysis and training.\\

{\verb|oov-off|} & {Turns off out-of-vocabulary treatment (default). For analysis and training.}\\

\verb|show-config| & {Shows current values of all the properties above.}\\
\end{tabular}
}
\label{tab:ref}
\end{table}

\newpage
\small
\begin{spacing}{0}
\setlength{\bibsep}{1ex}
\bibliographystyle{LI-like}
\bibliography{cem,references}

\begin{thebibliography}{45}
\newcommand{\enquote}[1]{``#1''}
\expandafter\ifx\csname natexlab\endcsname\relax\def\natexlab#1{#1}\fi
\expandafter\ifx\csname url\endcsname\relax
  \def\url#1{{\tt #1}}\fi
\expandafter\ifx\csname urlprefix\endcsname\relax\def\urlprefix{URL }\fi

\bibitem[{Abend et~al.(2017)Abend, Kwiatkowski, Smith, Goldwater, and
  Steedman}]{Aben:15a}
Abend, Omri, Tom Kwiatkowski, Nathaniel Smith, Sharon Goldwater, and Mark
  Steedman. 2017.
\newblock Bootstrapping language acquisition.
\newblock {\em Cognition\/}, 164:116--143.

\bibitem[{Ajdukiewicz(1935)}]{ajdu:35}
Ajdukiewicz, Kazimierz. 1935.
\newblock Die syntaktische {K}onnexit\"at. In {\em Polish Logic 1920--1939\/}.
\newblock ed.Storrs McCall,  207--231. Oxford: Oxford University Press.
\newblock Trans. from {\em Studia Philosophica}, 1, 1-27.

\bibitem[{Bach(1976)}]{bach:76}
Bach, Emmon. 1976.
\newblock An extension of {C}lassical {T}ransformational {G}rammar.
\newblock In {\em Problems in Linguistic Metatheory: Proceedings of the 1976
  Conference at {M}ichigan {S}tate {U}niversity\/},  183--224. Lansing, MI:
  Michigan State University.

\bibitem[{Baldridge(2002)}]{baldridge02}
Baldridge, Jason. 2002.
\newblock {\em Lexically Specified Derivational Control in {C}ombinatory
  {C}ategorial {G}rammar\/}.
\newblock Doctoral Dissertation, University of Edinburgh.

\bibitem[{Bar-Hillel(1953)}]{BarH:53}
Bar-Hillel, Yehoshua. 1953.
\newblock A quasi-arithmetical notation for syntactic description.
\newblock {\em Language\/}, 29:47--58.

\bibitem[{Bar-Hillel, Gaifman and Shamir(1960/1964)Bar-Hillel, Gaifman, and
  Shamir}]{BarH:60}
Bar-Hillel, Yehoshua, Chaim Gaifman, and Eliyahu Shamir. 1960/1964.
\newblock On categorial and phrase structure grammars. In {\em Language and
  Information\/}.
\newblock ed.Yehoshua Bar-Hillel,  99--115. Reading, MA: Addison-Wesley.

\bibitem[{Boz\c{s}ahin(2023)}]{bozs:22-jlli}
Boz\c{s}ahin, Cem. 2023.
\newblock Referentiality and configurationality in the idiom and the phrasal
  verb.
\newblock {\em Journal of Logic, Language and Information\/}, 32:175--207.

\bibitem[{Brown(1998)}]{brow:98}
Brown, Penelope. 1998.
\newblock Children's first verbs in {T}zeltal: Evidence for an early verb
  category.
\newblock {\em Linguistics\/}, 36:713--753.

\bibitem[{Brown(1973)}]{Brow:73}
Brown, Roger. 1973.
\newblock {\em A First Language: the Early Stages\/}.
\newblock Cambridge, MA: Harvard University Press.

\bibitem[{Cabay and Jackson(1976)}]{caba:76}
Cabay, S, and LW~Jackson. 1976.
\newblock A polynomial extrapolation method for finding limits and antilimits
  of vector sequences.
\newblock {\em {SIAM} Journal on Numerical Analysis\/}, 13:734--752.

\bibitem[{Crain and Thornton(1998)}]{Crai:98}
Crain, Stephen, and Rosalind Thornton. 1998.
\newblock {\em Investigations in Universal Grammar\/}.
\newblock Cambridge, MA: MIT Press.

\bibitem[{Eisner(1996)}]{eisner96}
Eisner, Jason. 1996.
\newblock Efficient normal-form parsing for {C}ombinatory {C}ategorial
  {G}rammar.
\newblock In {\em Proceedings of the 34th Annual Meeting of the {ACL}\/},
  79--86.

\bibitem[{Ellis(1938)}]{elli:38}
Ellis, Willis~D. 1938.
\newblock {\em A Source Book of Gestalt Psychology\/}.
\newblock London: Routledge and Kegan Paul.

\bibitem[{Gallier(2011)}]{gall:11}
Gallier, Jean. 2011.
\newblock {\em Discrete Mathematics\/}.
\newblock Springer.

\bibitem[{Gentner(1982)}]{gentner82}
Gentner, Dedre. 1982.
\newblock Why nouns are learned before verbs: Linguistic relativity versus
  natural partitioning. In {\em Language Development, vol.2: Language, Thought
  and Culture\/}.
\newblock ed.Stan A.~Kuczaj II,  301--334. Hillsdale, New Jersey: Lawrence
  Erlbaum.

\bibitem[{Grimshaw(1990)}]{grimshaw90}
Grimshaw, Jane. 1990.
\newblock {\em Argument Structure\/}.
\newblock Cambridge, MA: MIT Press.

\bibitem[{Hale and Keyser(2002)}]{halekeyser02}
Hale, Ken, and Samuel~Jay Keyser. 2002.
\newblock {\em Prolegomenon to a Theory of Argument Structure\/}.
\newblock Cambridge, MA: MIT Press.

\bibitem[{Hale(1983)}]{Hale:83}
Hale, Kenneth. 1983.
\newblock {W}arlpiri and the grammar of non-configurational languages.
\newblock {\em Natural Language and Linguistic Theory\/}, 1:5--47.

\bibitem[{Hoeksema and Janda(1988)}]{hoeksemajanda88}
Hoeksema, Jack, and Richard~D. Janda. 1988.
\newblock Implications of process-morphology for {C}ategorial {G}rammar. In
  {\em Categorial Grammars and Natural Language Structures\/}.
\newblock eds.Richard~T. Oehrle, Emmon Bach, and Deirdre Wheeler. Dordrecht: D.
  Reidel.

\bibitem[{Klein and Sag(1985)}]{Klei:85}
Klein, Ewan, and Ivan Sag. 1985.
\newblock Type-driven translation.
\newblock {\em Linguistics and Philosophy\/}, 8:163--201.

\bibitem[{Klein(1994)}]{klei:94}
Klein, Wolfgang. 1994.
\newblock {\em Time in Language\/}.
\newblock London: Routledge.

\bibitem[{Klein, Li and Hendriks(2000)Klein, Li, and Hendriks}]{klei:00}
Klein, Wolfgang, Ping Li, and Henriette Hendriks. 2000.
\newblock Aspect and assertion in {M}andarin {C}hinese.
\newblock {\em Natural Language and Linguistic Theory\/}, 18:723--770.

\bibitem[{Koffka(1936)}]{koff:36}
Koffka, Kurt. 1936.
\newblock {\em Principles of Gestalt Psychology\/}.
\newblock London: Kegan Paul.

\bibitem[{Lambek(1958)}]{lambek58}
Lambek, Joachim. 1958.
\newblock The mathematics of sentence structure.
\newblock {\em American Mathematical Monthly\/}, 65:154--170.

\bibitem[{{Mac Lane}(1971)}]{macl:71}
{Mac Lane}, Saunders. 1971.
\newblock {\em Categories for the Working Mathematician\/}.
\newblock Berlin/New York: Springer.

\bibitem[{MacWhinney(2000)}]{MacW:00}
MacWhinney, B. 2000.
\newblock {\em The {CHILDES} project: Tools for analyzing talk. Third
  edition\/}.
\newblock Mahwah, NJ: Lawrence Erlbaum Associates.

\bibitem[{Moggi(1988)}]{mogg:88}
Moggi, Eugenio. 1988.
\newblock {\em Computational Lambda-Calculus and Monads\/}.
\newblock LFCS, University of Edinburgh.

\bibitem[{Montague(1970)}]{Mont:70a}
Montague, Richard. 1970.
\newblock {English} as a formal language. In {\em Linguaggi nella Societ\`{a} e
  nella Technica\/}.
\newblock ed.Bruno Visentini,  189--224. Milan: Edizioni di Communit\`{a}.
\newblock Reprinted as \citealt{Mont:74}:188-221.

\bibitem[{Montague(1973)}]{Montague73}
Montague, Richard. 1973.
\newblock The proper treatment of quantification in ordinary {E}nglish. In {\em
  Approaches to Natural Language\/}.
\newblock eds.J.~Hintikka and P.~Suppes. Dordrecht: D. Reidel.

\bibitem[{Ross(1967)}]{ross67}
Ross, John~Robert. 1967.
\newblock {\em Constraints on Variables in Syntax\/}.
\newblock Doctoral Dissertation, MIT.
\newblock Published as {\em Infinite Syntax!}, Ablex, Norton, NJ, 1986.

\bibitem[{Sapir(1924/1949)}]{sapi:24}
Sapir, Edward. 1924/1949.
\newblock The grammarian and his language. In {\em Selected Writings of {Edward
  Sapir} in Language, Culture, and Personality\/}.
\newblock ed.David~G. Mandelbaum. Berkeley: University of California Press.
\newblock Originally published in \textit{American Mercury 1: (1924) 149--155}.

\bibitem[{Schmerling(1981)}]{Schm:81}
Schmerling, Susan. 1981.
\newblock The proper treatment of the relationship between syntax and
  phonology.
\newblock Paper presented at the 55th annual meeting of the LSA, San Antonio
  TX.

\bibitem[{Schmerling(2018)}]{schm:18}
Schmerling, Susan. 2018.
\newblock {\em Sound and Grammar: a Neo-{S}apirian Theory of Language\/}.
\newblock Leiden/Boston: Brill.

\bibitem[{Sch{\"o}nfinkel(1920/1924)}]{schonfinkel24}
Sch{\"o}nfinkel, Moses~Ilyich. 1920/1924.
\newblock On the building blocks of mathematical logic. In {\em From {F}rege to
  {G}{\"o}del\/}.
\newblock ed.Jan~{van} Heijenoort. Harvard University Press, 1967.
\newblock Prepared first for publication by H. Behmann in 1924.

\bibitem[{Steedman(2000)}]{steedman00}
Steedman, Mark. 2000.
\newblock {\em The Syntactic Process\/}.
\newblock Cambridge, MA: {MIT} Press.

\bibitem[{Steedman(2020)}]{stee:20}
Steedman, Mark. 2020.
\newblock A formal universal of natural language grammar.
\newblock {\em Language\/}, 96:618--660.

\bibitem[{Steedman and Baldridge(2011)}]{steedmanbaldridge-guide}
Steedman, Mark, and Jason Baldridge. 2011.
\newblock {C}ombinatory {C}ategorial {G}rammar. In {\em Non-transformational
  Syntax\/}.
\newblock eds.R.~Borsley and Kersti B{\"o}rjars,  181--224. Oxford: Blackwell.

\bibitem[{Swadesh(1938)}]{swad:38}
Swadesh, Morris. 1938.
\newblock Nootka internal syntax.
\newblock {\em International Journal of American Linguistics\/}, 9:77--102.

\bibitem[{Thomason(1974)}]{Mont:74}
ed.Thomason, Richmond. 1974.
\newblock {\em Formal Philosophy: Papers of {R}ichard {M}ontague\/}.
\newblock New Haven, CT: Yale University Press.

\bibitem[{Tomasello(1992)}]{Toma:92}
Tomasello, Michael. 1992.
\newblock {\em First Verbs: a Case Study in Early Grammatical Development\/}.
\newblock Cambridge: Cambridge University Press.

\bibitem[{Wertheimer(1924)}]{wert:24}
Wertheimer, Max. 1924.
\newblock Gestalt theory.
\newblock English translation published in \cite{elli:38}.

\bibitem[{Williams(1994)}]{Will:94}
Williams, Edwin. 1994.
\newblock {\em Thematic Structure in Syntax\/}.
\newblock Cambridge, MA: MIT Press.

\bibitem[{Wojdak(2000)}]{wojd:00}
Wojdak, Rachel. 2000.
\newblock Nuu-chah-nulth modification: Syntactic evidence against category
  neutrality.
\newblock In {\em Papers for the 35th International Conference on {S}alish and
  Neighbouring Languages ({UBCWPL})\/}, vol.~3,  269--281.

\bibitem[{Woolford(2006)}]{Woolford:06}
Woolford, Ellen. 2006.
\newblock Lexical case, inherent case, and argument structure.
\newblock {\em Linguistic Inquiry\/}, 37:111--130.

\bibitem[{Zettlemoyer and Collins(2005)}]{zettlemoyercollins05}
Zettlemoyer, Luke, and Michael Collins. 2005.
\newblock Learning to map sentences to logical form: Structured classification
  with probabilistic categorial grammars.
\newblock In {\em Proc. of the 21st Conf. on {U}ncertainty in {A}rtificial
  {I}ntelligence\/}. Edinburgh.

\end{thebibliography}
\end{spacing}
\end{document}